\documentclass[letterpaper]{article} 
\usepackage{aaai2026}  
\usepackage{times}  
\usepackage{helvet}  
\usepackage{courier}  
\usepackage[hyphens]{url}  
\usepackage{graphicx} 
\usepackage{natbib}  
\usepackage{caption} 
\usepackage{algorithm}
\usepackage{algorithmic}

\usepackage{amsthm, amsmath, amssymb}
\usepackage{mathrsfs}
\usepackage{multirow}
\usepackage{booktabs}

\usepackage{newfloat}
\usepackage{listings}
\DeclareCaptionStyle{ruled}{labelfont=normalfont,labelsep=colon,strut=off} 
\floatstyle{ruled}
\newfloat{listing}{tb}{lst}{}
\floatname{listing}{Listing}
\title{X-ReID: Multi-granularity Information Interaction for Video-Based Visible-Infrared Person Re-Identification}
\author {
	Chenyang Yu\textsuperscript{\rm 1},
	Xuehu Liu\textsuperscript{\rm 2},
	Pingping Zhang\textsuperscript{\rm 3,\rm 4}\thanks{Corresponding author.},
	Huchuan Lu\textsuperscript{\rm 1}
	\\
}
\affiliations{
	\textsuperscript{\rm 1} School of Information and Communication Engineering, Dalian University of Technology, Dalian, China\\
	\textsuperscript{\rm 2} School of Computer Science and Artificial Intelligence, Wuhan University of Technology, Wuhan, China\\
	\textsuperscript{\rm 3} School of Future Technology, Dalian University of Technology, Dalian, China\\
	\textsuperscript{\rm 4} National Engineering Laboratory for Integrated Aero-Space-Ground-Ocean Big Data Application Technology, Xi'an, China\\
	asuradayuci@gmail.com;liuxuehu@whut.edu.cn;{\{zhpp, lhchuan\}@dlut.edu.cn}
}

\begin{document}
\maketitle
\begin{abstract}
Large-scale vision-language models (e.g., CLIP) have recently achieved remarkable performance in retrieval tasks, yet their potential for Video-based Visible-Infrared Person Re-Identification (VVI-ReID) remains largely unexplored.
The primary challenges are narrowing the modality gap and leveraging spatiotemporal information in video sequences.
To address the above issues, in this paper, we propose a novel cross-modality feature learning framework named X-ReID for VVI-ReID.
Specifically, we first propose a Cross-modality Prototype Collaboration (CPC) to align and integrate features from different modalities, guiding the network to reduce the modality discrepancy.
Then, a Multi-granularity Information Interaction (MII) is designed, incorporating short-term interactions from adjacent frames, long-term cross-frame information fusion, and cross-modality feature alignment to enhance temporal modeling and further reduce modality gaps.
Finally, by integrating multi-granularity information, a robust sequence-level representation is achieved.
Extensive experiments on two large-scale VVI-ReID benchmarks (\emph{i.e.}, HITSZ-VCM and BUPTCampus) demonstrate the superiority of our method over state-of-the-art methods.
The source code is released at https://github.com/AsuradaYuci/X-ReID.
\end{abstract}
\begin{figure}[h]
	\centering
	\includegraphics[width=0.98\columnwidth]{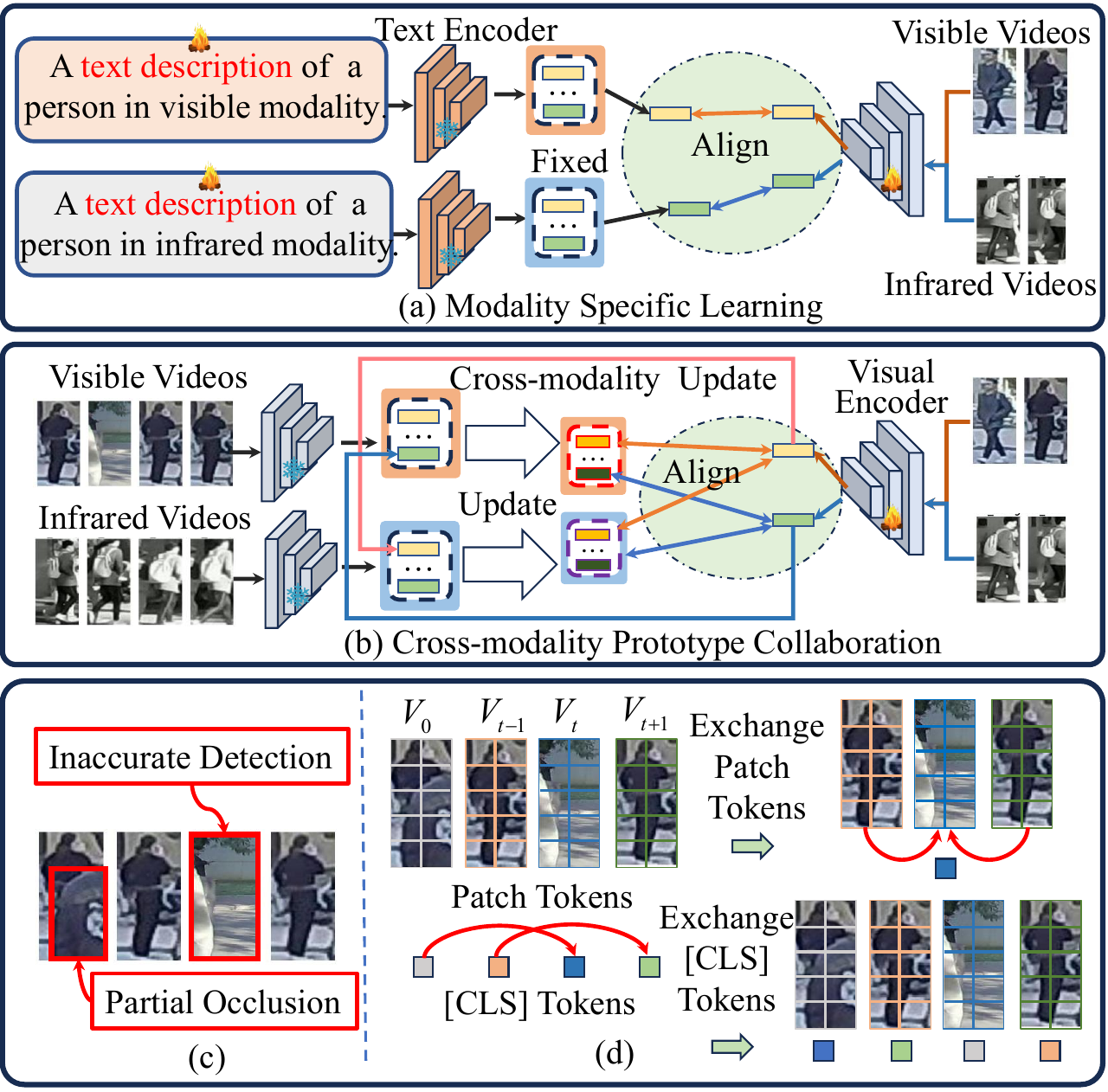}
	\caption{Illustration of our motivations.
	}
	\label{fig:intro}
\end{figure}
\section{Introduction}
Video-based Visible-Infrared person Re-IDentification (VVI-ReID) aims to retrieve video sequences of the same person captured by non-overlapping cameras operating in different modalities.
Over the past decade, various image-based VI-ReID methods~\cite{ye2021deep,lu2023learning,zhang2023diverse} have been developed.
However, image-based person ReID~\cite{zhang2021hat,yu2025hierarchical,wang2025idea,wang2025decoupled} is highly dependent on the quality of static images, making it sensitive to noise, viewpoint changes, and other variations.
In contrast, videos offer more comprehensive visual and motion information, providing valuable cues against these challenges.
As a result, VVI-ReID is gaining increasing attention.

Technically, there are two main challenges in addressing VVI-ReID: (1) bridging the gap between visible and infrared videos and (2) extracting robust and discriminative temporal features.
To reduce the modality gap, the common practice is to introduce additional auxiliary information, such as body shape features~\cite{feng2023shape} and anaglyph images~\cite{li2023intermediary}.
However, it is often a tedious process to generate such auxiliary information for each sample.
Recently, the large-scale vision-language model CLIP~\cite{radford2021learning} has achieved remarkable success in visible person ReID.
However, the application of CLIP to VVI-ReID has not been thoroughly investigated.
A major challenge is that, compared with general ReID methods like CLIP-ReID~\cite{li2023clip}, TF-CLIP~\cite{yu2024tf} and CLIMB-ReID~\cite{yu2025climb}, it is difficult to directly learn modality-shared textual prompts for VVI-ReID.
As shown in Fig.~\ref{fig:intro} (a), a straightforward solution is to learn label-specific textual prompts for each modality.
However, this not only requires an extra training phase but also fails to reduce the inherent modality gap.

On the other hand, harnessing spatiotemporal information is essential for VVI-ReID.
Some preliminary works~\cite{liu2021watching,hou2024three,lin2019tsm} extract appearance features from individual frames and then integrate them into a video-level representation using a temporal pooling layer or an LSTM~\cite{schmidhuber1997long}.
As shown in Fig.~\ref{fig:intro} (c), partial occlusions and inaccurate detection can corrupt the learned features, leading to substantial performance degradation.
Lately, various methods~\cite{liu2024video,wu2022cavit} considering both spatial and temporal representations, have achieved superior performance.
However, these methods overlook the reduction of modality discrepancy during temporal modeling.

To address the above issues, we propose a novel framework named X-ReID for VVI-ReID.
It is composed of two key components: Cross-modality Prototype Collaboration (CPC) and Multi-granularity Information Interaction (MII).
Technically, CPC efficiently transfers knowledge from CLIP and reduces the modality gap.
Taking visible modality as an example, as shown in Fig.~\ref{fig:intro} (b), we leverage the pre-trained CLIP visual encoder to initialize CLIP prototypes for each person.
Then, we select cross-modal infrared samples from the training batch to update these prototypes.
In this way, visible modal prototypes that fuse information from two modalities is obtained.
Similarly, we can get the updated infrared modal prototypes.
Subsequently, these refined prototypes are used to collaboratively guide the training of the visual encoder, thereby enhancing representation learning for the VVI-ReID task.
Meanwhile, we further design a MII to efficiently capture multigranular temporal information in videos.
As shown in Fig.~\ref{fig:intro} (d), we exchange the whole patch token features back-and-forth across adjacent frames along the channel dimension to reconstruct the current frame.
Then, a Short-term Information Interaction (SII) is designed to capture short-term temporal information in adjacent frames.
In addition, a Long-term Information Interaction (LII) is developed by exchanging [CLS] tokens over longer temporal intervals to capture long-term information across frames.
Furthermore, a Cross-Modality Information Interaction (CII) is proposed to reduce the modality gap between visible and infrared video-level features.
Extensive experiments on two VVI-ReID benchmarks clearly demonstrate the effectiveness of our method.

Our contributions can be summarized as follows:
\begin{itemize}
	\item We propose a novel framework named X-ReID for VVI-ReID. A Cross-modality Prototype Collaboration (CPC) is designed to efficiently transfer CLIP's knowledge and reduce the modality discrepancy.
	\item We propose a Multi-granularity Information Interaction (MII) to efficiently capture multi-granularity temporal information from videos while narrowing the modality gap through cross-modality information interaction.
	\item Extensive experiments demonstrate that our X-ReID outperforms state-of-the-art VVI-ReID methods on two benchmarks, \emph{i.e.}, HITSZ-VCM and BUPTCampus.
\end{itemize}
\section{Related Works}
\label{sec:intro}
\begin{figure*}[t]
	\centering
	\includegraphics[width=0.98\textwidth]{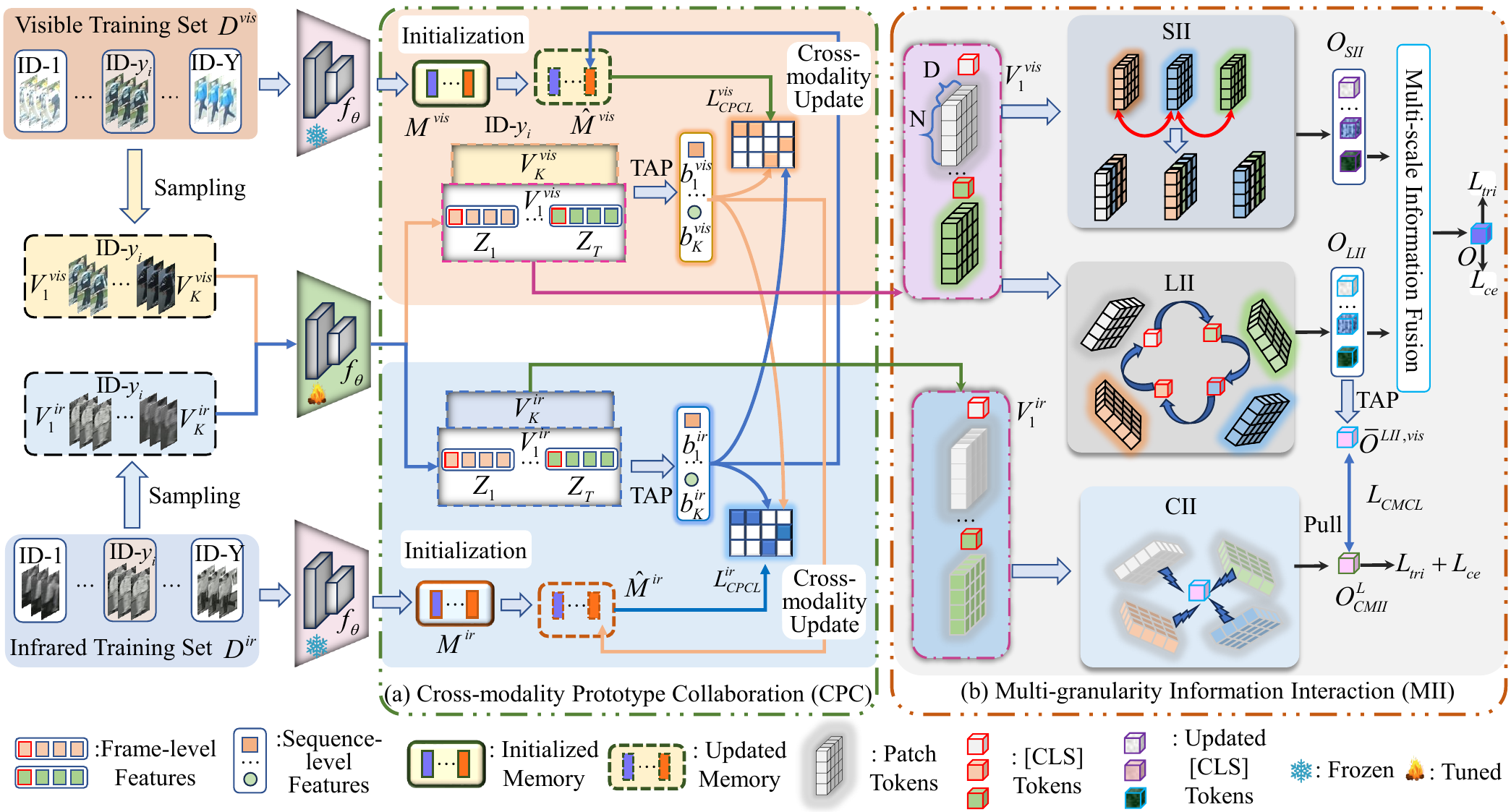}
	\caption{Illustration of the proposed X-ReID framework.
	}
	\label{fig:overall}
\end{figure*}
\subsection{Person ReID with CLIP}
CLIP has achieved remarkable success in various multi-modal understanding tasks~\cite{pang2024open,gong2025devil}.
Recently, several studies have extended CLIP to person ReID.
To overcome the problem of missing text labels in ReID tasks, Li \emph{et al.}~\cite{li2023clip} design a prompt learning strategy to generate the label-specific text features for image-based person ReID.
Yu \emph{et al.}~\cite{yu2025clip} construct modality-shared textual prompts to reduce the modality gap for visible-infrared person ReID.
Wang \emph{et al.}~\cite{wang2025makes} introduce attribute prompt composition for domain generalization ReID.
Although the above methods have achieved great success, they inevitably introduce additional training stages or require additional prompt networks.
To address these issues, Li \emph{et al.}~\cite{li2023prototypical} propose a prompt-free framework that only utilizes visual features for fine-tuning.
Yu \emph{et al.}~\cite{yu2024tf} propose a text-free CLIP framework that extracts identity-specific prototypes as a substitute for text features.
Yu \emph{et al.}~\cite{yu2025climb} propose a hybrid CLIP-Mamba framework for robust person ReID.
Nonetheless, the vast potential of CLIP in advancing the learning of modality-invariant features for VVI-ReID remains under-explored.
Inspired by the above works, in this paper, we design the CPC to efficiently transfer CLIP's knowledge and reduce the modality gap.
\subsection{Video-based Visible-Infrared Person ReID}
VVI-ReID is attracting growing attention, as videos provide more comprehensive spatial-temporal information than static images.
The release of large-scale datasets, such as HITSZ-VCM~\cite{lin2022learning} and BUPTCampus~\cite{du2023video}, has also prompted a gradual shift from VI-ReID to VVI-ReID.
For example, Lin \emph{et al.}~\cite{lin2022learning} propose to use adversarial learning~\cite{UniMRSeg} to achieve modality-invariant features based on additional modality labels.
Li \emph{et al.}~\cite{li2023intermediary} introduce anaglyph images as an intermediary to bridge different modalities and guide the model in learning modality-independent features.
They all use LSTM~\cite{schmidhuber1997long} to capture temporal cues between frames.
Zhou \emph{et al.}~\cite{zhou2023video} further utilize graph networks to explore the cross-view and cross-modal correlations.
Unfortunately, these methods exhibit suboptimal performance due to the inherent limitations of CNN in capturing global features and the inability of LSTM to effectively model long-range dependencies in long sequences.
To address above issues, Feng \emph{et al.}~\cite{feng2024cross} introduce a Transformer structure to explore global feature relationships within frames and temporal cues across frames.
Despite some success, these approaches overlook the reduction of modality discrepancy during temporal modeling.
In this paper, we propose a cross-modality information interaction mechanism to explicitly constrain the feature discrepancy between modalities.
\section{Proposed Method}
As illustrated in Fig.~\ref{fig:overall}, our proposed framework includes two components: Cross-modality Prototype Collaboration (CPC) and Multi-granularity Information Interaction (MII).
\subsection{Preliminaries}
For VVI-ReID, let $vis$ and $ir$ represent the visible modality and infrared modality, respectively.
Thus, the training set can be denoted as $\{D^{vis},D^{ir},Y\}$, where $D^{vis}$=$\{V_{i}^{vis}\}_{i=1}^{N_{vis}}$ represents $N_{vis}$ visible videos, $D^{ir}$=$\{V_{i}^{ir}\}_{i=1}^{N_{ir}}$ represents $N_{ir}$ infrared videos, and $Y$=$\{y_i\}_{i=1}^{N_{c}}$ is the corresponding label set.
Since the processing for the two modalities is similar, we focus on describing the detailed process of the visible modality for simplicity.
Taken a visible video sequence $V^{vis}$=$\{I_t^{vis}\}_{t=1}^{T}$ containing $T$ frames as an example, $I_t^{vis}\in \mathbb{R}^{H\times W\times 3}$ is the $t$-th frame, where $H$ and $W$ represent the number of height and width, respectively.
As shown in Fig.~\ref{fig:overall}, the CLIP visual encoder $f_{\theta}$ is adopted to extract high-level representations.
Each frame is first divided into $N$ patches and processed by $L$ Transformer layers, yielding a frame-level feature
$Z_t = \{ z_{t}^{cls}; z_{t}^{1}; z_{t}^{2}; \dots; z_{t}^{N} \} \in \mathbb{R}^{(1+N) \times D}$.
Here, $z_{t}^{cls}$ and $z_{t}^{n}$ denote the $D$-dimensional [CLS] and patch tokens, respectively.
The [CLS] token $z_{t}^{cls}$ is then projected into a unified visual-language embedding space via a linear layer, producing $v_{t}^{vis} \in \mathbb{R}^{1 \times d}$.
Finally, a Temporal Average Pooling (TAP) aggregates all frame-level features to obtain the sequence-level feature $b_{y_i}^{vis}$.
\subsection{Cross-modality Prototype Collaboration}
To fully harness the potential of CLIP for VVI-ReID, we first address the issue of missing text labels.
A common solution~\cite{chen2023unveiling} is to learn modality-specific textual prompts, such as “A visible/infrared video of a XXXX person”.
However, this solution not only requires an additional training phase but also leads to suboptimal results.
Although TF-CLIP~\cite{yu2024tf} further propose a text-free solution, it still fail to reduce the gap between different modalities.
To solve the above problems, we propose the Cross-modality Prototype Collaboration (CPC).

\textbf{Memory Initialization.} Inspired by \cite{yu2024tf,li2023prototypical}, we propose to transfer the knowledge of CLIP to VVI-ReID by utilizing a pre-trained visual encoder to extract identity-specific prototypes, rather than textual prompts.
As shown in Fig.~\ref{fig:overall} (a), we traverse the training sets $D^{vis}$ and $D^{ir}$ separately by the pre-trained $f_{\theta}$ to construct modality-specific memories $M^{vis}$ and $M^{ir}$.

Taking visible modality as an example, we maintain a prototype in $M^{vis}$ for each identity.
Specifically, once all the features belonging to the identity $y_i$ are obtained, the average of them can represent the identity-specific feature $M_{y_{i}}^{vis}=\frac{1}{N_i}\sum_{b\in y_i}b$ to initialize each prototype in the memory.
Here, $N_i$ is the number of videos belonging to the identity $y_i$.
Therefore, the initialized memory $M^{vis}\in \mathbb{R}^{Y\times d}$ has $Y$ entries, in which $d$ represent the dimension of the features.

\textbf{Cross-modality Update.} Notice that when the parameters of CLIP's visual encoder are updated, the prototype of label $y_i$ will also be updated by:
\begin{equation}\label{updated}
M_{y_i}^{vis}\leftarrow \mu \cdot M_{y_i}^{vis}  + (1-\mu) \cdot b_{y_i}^{*},
\end{equation}
where $\mu$ is the momentum factor. $b^{*}$ represents the samples selected from current training batch $B$.
Typically, only modality-specific updates are performed.
However, conducting contrastive learning solely within each modality does not meet the requirements of VVI-ReID.
To reduce the modality gap, as shown in Fig.~\ref{fig:overall} (a), we propose to utilize different modality samples $b^{ir}$ from the current training batch $B$ to update the memories.
Specifically, for the prototype of label $y_i$, $b^{*}$ can be obtained in the following ways:
\begin{equation}\label{hard}
	b_{y_i}^{ir}\leftarrow \mathop{\arg\min}\limits_{b^{ir}} b^{ir}\cdot M_{y_i}^{vis}, b^{ir}\in B_{y_i}^{ir},
\end{equation}
where $B_{y_i}^{ir}$ is the feature set of infrared samples with label $y_i$ in current mini-batch.
$b_{y_i}^{ir}$ is the feature of the hard sample which has the lowest similarity with $M_{y_i}^{vis}$.
Thanks to the aforementioned cross-modality update strategy, we can obtain a fused modality prototype $\hat{M}_{y_i}^{vis}$.
Similarly, we can get another fused modality prototype $\hat{M}_{y_i}^{ir}$.

\textbf{Cross-modality Collaboration.} In each batch, we randomly sample $K$ videos for each modality from $P$ persons.
In fact, forcing samples to simultaneously approximate two prototypes that fuse information from two modalities helps learn modality-independent features.
Thus, as shown in Fig.~\ref{fig:overall} (a), we further propose a Cross-modality Prototype Collaboration Loss (CPCL) to achieve the cooperation of cross-modality memories,
\begin{equation}\label{loss1}
L_{CPCL}^{vis}=-log\frac{\exp(<b_{y_i}^{*}\cdot \hat{M}_{y_i}^{vis}>)}{\sum_{j=1}^{P}\exp(<b_{y_i}^{*} \cdot \hat{M}_{j}^{vis}>)},
\end{equation}
\begin{equation}\label{loss2}
L_{CPCL}^{ir}=-log\frac{\exp(<b_{y_i}^{*}\cdot \hat{M}_{y_i}^{ir}>)}{\sum_{j=1}^{P}\exp(<b_{y_i}^{*} \cdot \hat{M}_{j}^{ir}>)},
\end{equation}
where $\hat{M}_{y_i}^{vis}$ and $\hat{M}_{y_i}^{ir}$ are the prototypes of identity $y_i$.
$<\cdot>$ is the cosine similarity function.
$*$ means that all samples of the two modalities participate in.
By concurrently aligning samples from two modalities with prototypes that integrate cross‑modal information, the modality gap can be narrowed, leading to more robust features.
\subsection{Multi-granularity Information Interaction}
To further capture spatiotemporal information, we propose a Multi-granularity Information Interaction (MII).
As shown in Fig.~\ref{fig:overall} (b), it is mainly composed of Short-term Information Interaction (SII), Long-term Information Interaction (LII) and Cross-modality Information Interaction (CII).

\textbf{Short-term Information Interaction.} To capture short-term temporal information, we introduce the SII across adjacent frames.
The key idea is exchanging patch tokens of adjacent frames to reconstruct the current frame and interacting with [CLS] tokens to capture short-term information.

\begin{figure}[t]
	\centering
	\includegraphics[width=1.0\linewidth]{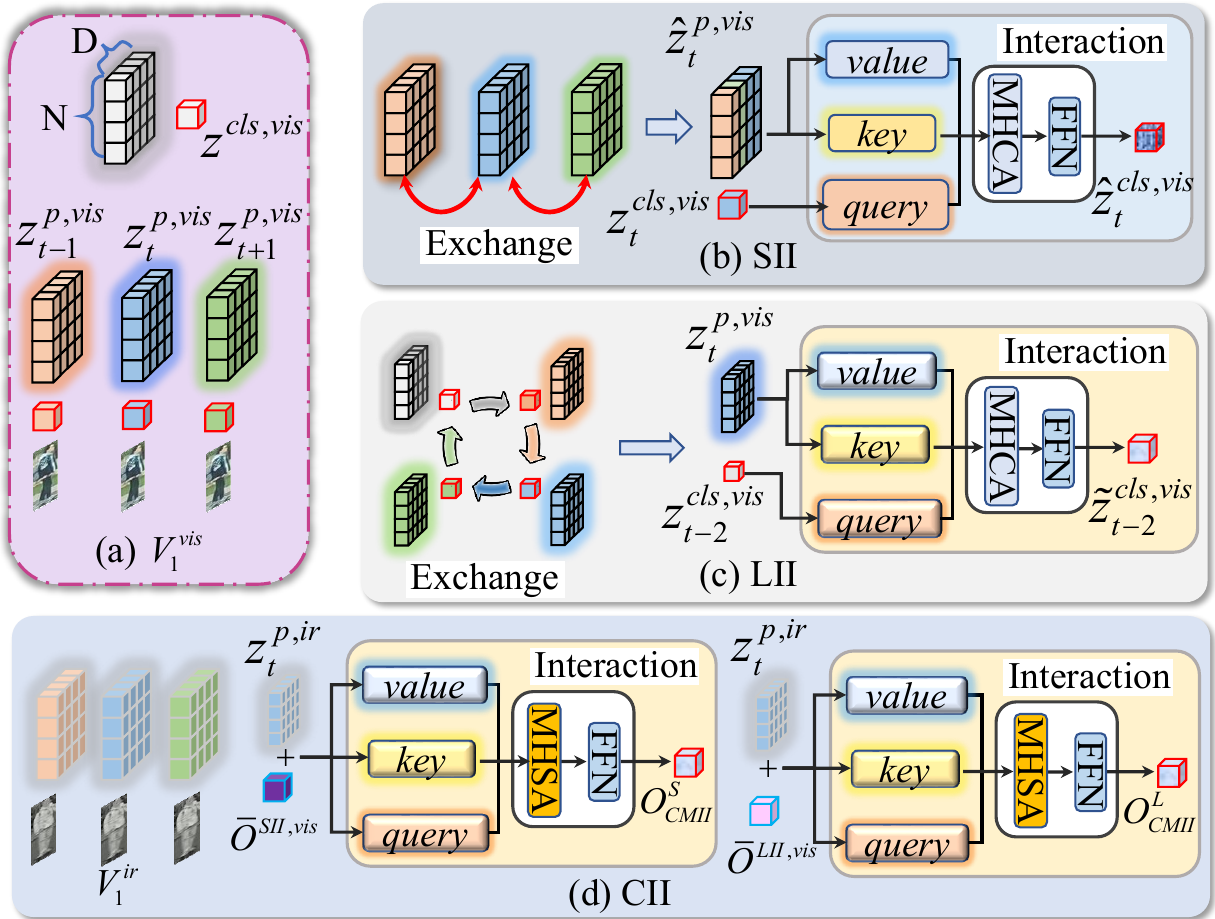}
	\caption{Illustration of our MII.
	}
	\label{fig:CE}
\end{figure}

As shown in Fig.~\ref{fig:CE} (a), we take a visible video $V_1^{vis}=\{[z_t^{cls,vis}, z_t^{p,vis}]\}_{t=1}^{T}$ as an example.
For the $t$-th frame, its patch tokens $z_t^{p,vis}\in \mathbb{R}^{N\times D}$ are exchanged with two adjacent frames in the channel dimension to obtain the reconstructed patch tokens $\hat{z}_t^{p,vis}$.
As shown in Fig.~\ref{fig:CE} (b), the above operation can be defined as:
\begin{equation}
\label{exchange}
\small
\hat{z}_t^{p,vis}=[z_{t-1}^{p,vis}(1:\frac{D}{4});z_{t+1}^{p,vis}(\frac{D}{4}+1:\frac{D}{2});z_{t}^{p,vis}(\frac{D}{2}+1:D)],
\end{equation}
where $(m:n)$ are the channel indexes from $m$ to $n$.
$[;]$ is the concatenation along the channel dimension.
It is worth noting that the first and last frames lack an adjacent frame, and we simply copy the current frame to perform the exchange operation.
SII further takes the reconstructed patch tokens $\hat{z}_t^{p,vis}$ and the original [CLS] token $z_t^{cls,vis}$ as inputs to capture short-term information.
As shown in Fig.~\ref{fig:CE} (b), $z_t^{cls,vis}$ is used as $query$ and $\hat{z}_t^{p,vis}$ is employed as $key$ and $value$.
Then, a Multi-Head Cross-Attention (MHCA) followed by a Feed-Forward Network (FFN)~\cite{vaswani2017attention} is utilized to capture short-term information:
\begin{equation}\label{mhcs}
	\hat{z}_t^{cls,vis}=FFN(MHCA(query,key,value)).
\end{equation}
In this way, the [CLS] token of the current frame can effectively capture short-term temporal dependencies by leveraging information from adjacent frames.
Finally, we can obtain $O^{SII,vis}$=$\{\hat{z}_t^{cls,vis}\}_{t=1}^{T}$ for the input sequence.

\textbf{Long-term Information Interaction.} Considering that SII can only capture short-term information in adjacent frames, we further propose a LII for long-term modeling.

As shown in Fig.~\ref{fig:CE} (c), we exchange [CLS] tokens in sequences with longer time steps.
Taking step $S=2$ as an example, the [CLS] token of the current frame $z_{t-2}^{cls,vis}$ will be matched with the patch tokens ${z}_t^{p,vis}$ of other frames.
Similar to SII, we also perform an interaction between $z_{t-2}^{cls,vis}$ and ${z}_t^{p,vis}$.
As shown in Fig.~\ref{fig:CE} (c), $z_{t-2}^{cls,vis}$ is used as $query$ and ${z}_t^{p,vis}$ is employed as $key$ and $value$.
The interaction of the LII is shown as follows:
\begin{equation}\label{mhcs2}
\widetilde{z}_{t-2}^{cls,vis}=FFN(MHCA(query,key,value)),
\end{equation}
where $\widetilde{z}_{t-2}^{cls,vis}$ is the output of LII which contains the long-term information.
Afterwards, we can obtain $O^{LII,vis}$=$\{\widetilde{z}_t^{cls,vis}\}_{t=1}^{T}$ for the input sequence.
As shown in Fig.~\ref{fig:overall} (b), we finally aggregate the outputs of two branches to obtain the final output:
\begin{equation}
O^{vis}=Mean(O^{SII,vis}+O^{LII,vis}),
\end{equation}
where $Mean$ is the average operation used to fuse short-term and long-term information.
With the same operations, the infrared feature ${O}^{ir}$ can also be obtained.
\begin{table*}
	\begin{center}
		\resizebox{0.90\textwidth}{!}
		{
			\begin{tabular}{ccccccccc}
				\hline
				\multirow{2}{*}{Method}         & \multicolumn{4}{c}{Infrared to Visible (I2V)}                & \multicolumn{4}{c}{Visible to Infrared (V2I)}                 \\
				  & Rank-1        & Rank-5        & Rank-20       & mAP           & Rank-1        & Rank-5        & Rank-20       & mAP           \\ \hline
				DDAG~\cite{ye2020dynamic}                      & 54.6          & 69.8          & 81.5          & 39.3          & 59.0          & 74.6          & 84.0          & 41.5          \\
				LBA~\cite{park2021learning}                    & 46.4          & 65.3          & 79.4          & 30.7          & 49.3          & 69.3          & 82.2          & 32.4          \\
				MPANet~\cite{wu2021discover}                   & 46.5          & 63.1          & 77.8          & 35.3          & 50.3          & 67.3          & 79.7          & 37.8          \\
				VSD~\cite{tian2021farewell}                    & 54.5          & 70.0          & 82.0          & 41.2          & 57.5          & 73.7          & 83.6          & 43.5          \\
				CAJL~\cite{ye2021channel}                      & 56.6          & 73.5          & 84.1          & 41.5          & 60.1          & 74.6          & 84.5          & 42.8          \\
				MITML~\cite{lin2022learning}                  & 63.7          & 76.8          & 86.3          & 45.3          & 64.5          & 79.0          & 87.1          & 47.7          \\
				IBAN~\cite{li2023intermediary}                & 65.0          & 78.3          & 87.2          & 48.8          & 69.6          & 81.5          & 88.8          & 51.0          \\
				SAADG~\cite{zhou2023video}                      & 69.2          & 80.6          & 88.7          & 53.8          & 73.1          & 83.5          & 89.7          & 56.1          \\
				CLIP-ReID~\cite{li2023clip}                                     & 54.5          & 70.7          & 82.8          & 40.0          & 56.9          & 73.7          & 84.8          & 41.4          \\
				PCL~\cite{li2023prototypical}                                           & 64.4          & 76.7          & 86.1          & 51.3          & 67.0          & 79.6          & 88.1          & 53.7          \\
				TF-CLIP~\cite{yu2024tf}                                        & 61.9          & 74.7          & 85.0          & 49.0          & 64.8          & 77.7          & 86.8          & 51.0          \\
				CST~\cite{feng2024cross}                       & 69.4          & 81.1          & \underline{89.7}          & 51.2          & 72.6          & 83.4          & 89.8          & 53.0          \\
				DIRL~\cite{wang2025dirl}                      & 65.2         & 79.1         & -          & 47.9          & 67.0          & 81.7         & -          & 50.2          \\
				HD-GI~\cite{zhou2025hierarchical}                      & \underline{71.4}          & \underline{81.7}          & 84.9          & \underline{58.0}          & \underline{75.0}          & \underline{84.4}          & \underline{89.9}          & \underline{60.2}          \\
				\hline
				Ours                               & \textbf{73.4} & \textbf{85.0} & \textbf{92.3} & \textbf{60.5} & \textbf{76.1} & \textbf{87.1} & \textbf{93.7} & \textbf{59.6} \\ \hline
			\end{tabular}
		}
		\caption{Performance comparison on HITSZ-VCM.}
		\label{table:SOTA1}
	\end{center}
\end{table*}

\textbf{Cross-Modality Information Interaction.} In VVI-ReID, it is essential to reduce the discrepancy between the two modalities.
Although the above modules are effective in capturing temporal information, they do not address the significant gap between visible and infrared modalities.
To mitigate this issue, as illustrated in Fig.~\ref{fig:CE} (d), a CII module is further proposed to reduce the modality discrepancy.
Specifically, given the output of LII for the visible modality ${O}^{LII,vis}$, and the feature of the same person in another modality ${V}_1^{ir}$= $\{[{z}_t^{cls,ir}, {z}_t^{p,ir}]\}_{t=1}^{T}$.
The sequence-level feature $\bar{O}^{LII,vis}$ of the visible modality is first obtained via a TAP over ${O}^{LII,vis}$.
Then it concatenated along the channel dimension with the patch tokens of the $t$-th frame ${z}_t^{p,ir}$ $\in$ $\mathbb{R}^{N\times D}$ from the infrared modality, resulting in the input to CII: $[\bar{{O}}^{LII,vis}, {z}_t^{p,ir}]$ $\in$ $\mathbb{R}^{(N+1)\times D}$.
As shown in Fig.~\ref{fig:CE} (d), the concatenated feature is further processed by a MHSA followed by a FFN to enable cross-modality feature interaction, yielding the output $[\hat{{O}}^{LII,vis}, \hat{{z}}_t^{p,ir}]$ $\in$ $\mathbb{R}^{(N+1)\times D}$.

The frame-level feature after interaction is represented by $\hat{O}^{LII,vis}$, and all frame-level features are further aggregated via a TAP to obtain the final sequence-level representation ${O}_{CMII}^{L}$ incorporating cross-modal information.
Similarly, when the input to the CII module originates from the SII, the resulting sequence-level feature can be denoted by ${O}_{CMII}^{S}$.
To explicitly minimize the gap between two modalities, as shown in Fig.~\ref{fig:overall} (b), a Cross-Modality Constraint Loss (CMCL) is further introduced.
Taking the output of LII, $\bar{O}^{LII,vis}$ and the corresponding output of CII as an example,
\begin{equation}\label{loss2}
	L_{CMCL} = \frac{1}{PK}\sum^{PK}(\|\bar{O}^{LII,vis}-{O}_{CMII}^{L} \|_{2}^{2}).
\end{equation}

By minimizing the distance between features from the current modality and their counterparts after cross-modality interaction, $L_{CMCL}$ effectively reduces the modality gap.
Similarly, the features obtained from the  SII are also subjected to the same constraint.
Distinct from existing methods, the proposed method not only emphasizes temporal modeling but also explicitly addresses modality discrepancy.
\subsection{Training and Inference}
During training, we employ four different losses: the cross-modality prototype collaboration loss $L_{CPCL}$, the triplet loss $L_{tri}$~\cite{hermans2017defense}, the cross-entropy loss $L_{ce}$ and the cross-modality constraint loss $L_{CMCL}$.
Finally, the overall loss $L_{total}$ is defined as:
\begin{equation}\label{loss2}
	L_{total}=L_{CPCL}+L_{tri}+L_{ce}+L_{CMCL}.
\end{equation}
During inference, the CII module is not involved.
And the original sequence-level feature and the multi-scale temporal feature $O$ are concatenated to obtain the final representation.
\begin{table*}
	\begin{center}
		\resizebox{0.90\textwidth}{!}
		{
			\begin{tabular}{ccccccccc}
				\hline
				\multirow{2}{*}{Method}      & \multicolumn{4}{c}{Infrared to Visible (I2V)}                & \multicolumn{4}{c}{Visible to Infrared (V2I)}                 \\
				           & Rank-1        & Rank-5        & Rank-20       & mAP           & Rank-1        & Rank-5        & Rank-20       & mAP           \\ \hline
				AlignGAN~\cite{wang2019rgb}                & 28.0          & 49.1          & 66.6          & 30.3          & 35.4          & 53.9          & 68.7          & 35.1          \\
				DDAG~\cite{ye2020dynamic}                   & 40.9          & 61.4          & 78.5          & 40.4          & 46.3          & 68.2          & 81.3          & 43.1          \\
				LBA~\cite{park2021learning}                  & 32.1          & 54.9          & 72.6          & 32.9          & 39.1          & 58.7          & 75.4          & 37.1          \\
				CAJL~\cite{ye2021channel}                   & 40.5          & 66.8          & 81.2          & 41.5          & 45.0          & 70.0          & 83.3          & 43.6          \\
				AGW~\cite{ye2021deep}                      & 36.4          & 60.1          & 76.5          & 37.4          & 43.7          & 64.4          & 80.0          & 41.1          \\
				MMN~\cite{ye2021deep}                          & 40.9          & 67.2          & 80.6          & 41.7          & 43.7          & 65.2          & 80.9          & 42.8          \\
				DART~\cite{yang2022learning}                 & 52.4          & 70.5          & 84.0          & 49.1          & 53.3          & 75.2          & 85.7          & 50.5          \\
				MITML~\cite{lin2022learning}                & 49.1          & 67.9          & 81.5          & 47.5          & 50.2          & 68.3          & 83.5          & 46.3          \\
				DEEN~\cite{zhang2023diverse}                & 49.8          & 71.6          & 85.8          & 48.6          & 53.7          & 74.8          & 87.6          & 50.4          \\
				SAADG~\cite{zhou2023video}                                         & 63.5          & 79.3          & 88.3          & 56.7          & 59.0          & 76.7          & 87.9          & 56.0          \\
				AuxNet~\cite{du2023video}                  & 66.5          & 83.1         & \underline{90.4}          & 64.1         & 65.2          & 81.8          & \underline{89.8}          & 62.2         \\
				CLIP-ReID~\cite{li2023clip}                                  & 39.2          & 62.1          & 82.6          & 39.8          & 41.3          & 63.5          & 78.5          & 40.8          \\
				PCL~\cite{li2023prototypical}          & 61.2          & 80.2          & 89.9          & 58.6          & 61.7          & 78.9          & 88.1          & 57.3          \\
				TF-CLIP~\cite{yu2024tf}                                      & 57.1          & 80.4          & 89.4          & 56.6          & 59.6          & 78.7          & 87.4          & 55.3          \\
				DIRL~\cite{wang2025dirl}                      & \underline{67.6}         & 83.2         & -          & 63.4          & 67.2          & 83.0         & -          & \underline{65.3}          \\
				VLD~\cite{li2025video}                                    & 66.7          & \underline{85.8}         & -          & \underline{66.0}          & \underline{67.4}         & \underline{84.2}         & -          & 64.1          \\
				\hline
				Ours                                    & \textbf{68.2} & \textbf{88.4} & \textbf{94.3} & \textbf{68.5} & \textbf{68.8} & \textbf{84.8} & \textbf{92.7} & \textbf{65.9} \\ \hline
			\end{tabular}
		}
		\caption{Performance comparison on BUPTCampus.}
		\label{table:SOTA2}
	\end{center}
\end{table*}
\section{Experiment}
\label{sec:exp}
\subsection{Datasets and Evaluation Protocols}
In this work, we evaluate our approach on two  VVI-ReID datasets, \emph{i.e.}, HITSZ-VCM~\cite{lin2022learning} and BUPTCampus~\cite{du2023video}.
More details of these datasets can be found in the corresponding references.
The testing protocol contains both Infrared-to-Visible (I2V) and Visible-to-Infrared (V2I).
Following common practices, we adopt the Cumulative Matching Characteristic (CMC) and mean Average Precision (mAP) to measure the performance.
\subsection{Implementation Details}
Our model is trained with one NVIDIA A100 GPU (80G memory) and implemented with the PyTorch toolbox.
We use the ViT-B/16 from CLIP~\cite{radford2021learning} as the feature encoder.
During training, we adopt random flipping, random cropping and random erasing~\cite{zhong2020random} for data augmentation.
We set $\mu$ to $0.2$.
Following the existing VVI-ReID methods~\cite{zhou2023video,feng2024cross}, each frame is resized to $288\times144$.
We train the framework for 60 epochs in total.
The mini-batch size is 8, consisting of 4 identities, 2 video clips for each modality and 10 frames from each clip.
We utilize the Adam optimizer with the learning rate of 5 $\times 10^{-6}$.
Following the common practice\cite{yu2024tf}, we also warm up the model with 10 epochs, linearly increasing the learning rate from 5 $\times 10^{-7}$ to 5 $\times 10^{-6}$.
Afterwards, the learning rate is reduced by a factor of 0.1 at the 30th and 50th epochs.
\subsection{Comparison with State-of-the-art Methods}
In this section, we compare our method with other methods on two large-scale VVI-ReID benchmarks.
Experimental results are reported in Tab.~\ref{table:SOTA1} and Tab.~\ref{table:SOTA2}.
On the HITSZ-VCM and BUPTCampus datasets, our method achieves the best results of 73.4\%, and 68.2\% in Rank-1 accuracy under the I2V setting, respectively.
Compared with other VI-ReID methods~\cite{park2021learning,ye2021channel}, our method significantly outperforms them.
For example, our method achieves 68.2\% /68.8\% Rank-1 accuracy on BUPTCampus under the I2V and V2I settings, respectively, which surpasses DDAG~\cite{ye2020dynamic} by 27.3\% and 22.5\%.
The reason is that VI-ReID methods focus on the extraction of frame-level appearance information and do not consider modeling of temporal information within the videos.

As a representative VVI-ReID method, IBAN~\cite{li2023intermediary} introduces anaglyph images to bridge modality differences.
We do not introduce additional auxiliary training samples.
However, our method achieves 73.4\% Rank-1 accuracy on HITSZ-VCM under the I2V setting, which surpasses IBAN by 8.4\%.
In addition, CST~\cite{feng2024cross} uses Transformers to establish long-range temporal dependencies.
Our method obtains a higher Rank-1 by 3.5\% than CST under the V2I setting on HITSZ-VCM.
We attribute this to the fact that cross-modality information interaction helps to reduce the discrepancy between modalities.

Meanwhile, we further analyze and compare CLIP-based methods~\cite{li2023clip,li2023prototypical,yu2024tf}.
Although these methods have achieved great success in general ReID, their performance on VVI-ReID is far from ideal.
This is primarily because simple unimodal learning fails to effectively diminish the modality difference between visible and infrared.
Instead, we design the CPC to efficiently transfer knowledge from CLIP and reduce the modality gap.
As a result, our method achieves 68.8\% Rank-1 accuracy on BUPTCampus under the V2I setting, which surpasses VLD by 1.4\%.
These comparisons clearly demonstrate the superiority and effectiveness of the proposed
method on large-scale VVI-ReID datasets.
\subsection{Ablation Study}
We conduct ablation experiments on HITSZ-VCM and BUPTCampus datasets to assess the impact of different components.
In this subsection, we adopt PCL as the baseline.
The compared results are shown in Tab.~\ref{table:abla1}.
\begin{table}[t]
	\begin{center}
		\resizebox{1\linewidth}{!}
		{
			\begin{tabular}{ccccccccc}
				\hline
				\multirow{3}{*}{Method} & \multicolumn{4}{c}{HITSZ-VCM}                                & \multicolumn{4}{c}{BUPTCampus}                               \\
				& \multicolumn{2}{c}{I2V}           & \multicolumn{2}{c}{V2I} & \multicolumn{2}{c}{I2V}           & \multicolumn{2}{c}{V2I} \\
				& R-1 & \multicolumn{1}{c}{mAP}  & R-1       & mAP       & R-1 & \multicolumn{1}{c}{mAP}  & R-1      & mAP       \\ \hline
				Baseline                & 64.4   & \multicolumn{1}{c}{51.3} & 67.0         & 53.7      & 61.2   & \multicolumn{1}{c}{58.6} & 61.7        & 57.3      \\
				+ CPC                  & 67.7   & \multicolumn{1}{c}{57.1} & 71.4         & 56.6      & 63.6   & \multicolumn{1}{c}{62.6} & 64.4        & 61.2      \\
				+ MII                   & 73.4   & \multicolumn{1}{c}{60.5} & 76.1         & 59.6      & 68.2   & \multicolumn{1}{c}{68.5} & 68.8        & 65.9      \\ \hline
			\end{tabular}
		}
		\caption{Performance comparison of different components on HITSZ-VCM and BUPTCampus.}
		\label{table:abla1}
	\end{center}
\end{table}

\textbf{Effectiveness of CPC.}
As demonstrated in Tab.~\ref{table:abla1}, incorporating CPC into the baseline delivers an improvement of 4.8\% in mAP and 3.3\% in Rank-1 on HITSZ-VCM under the I2V setting.
Additionally, a significant improvement is also observed on BUPTCampus.
It is evident that our CPC effectively enhances performance across various metrics.
A plausible explanation for this improvement is that the proposed CPC can effectively transfer knowledge from CLIP and reduce the modality gap.
\begin{table}[t]
	\begin{center}
		\resizebox{0.90\linewidth}{!}
		{
			\begin{tabular}{ccccc}
				\hline
				\multirow{3}{*}{Ways} & \multicolumn{4}{c}{HITSZ-VCM}                                    \\
				& \multicolumn{2}{c}{I2V}           & \multicolumn{2}{c}{V2I} \\
				& R-1 & \multicolumn{1}{c}{mAP}  & R-1      & mAP       \\ \hline
				Vanilla                     & 53.9   & \multicolumn{1}{c}{44.0} & 55.9        & 42.0      \\
				Textual Prompts            & 54.5   & \multicolumn{1}{c}{40.0} & 56.9        & 41.4      \\
				Visual Memories            & 64.4   & \multicolumn{1}{c}{51.3} & 67.0        & 53.7      \\
				Cross-modal Update          & 67.5   & \multicolumn{1}{c}{55.4} & 70.5        & 54.7      \\
				Cross-modal Collaboration   & 61.9   & \multicolumn{1}{c}{49.0} & 64.8        & 51.0      \\
				CPC                     & 67.7   & \multicolumn{1}{c}{57.1} & 71.4        & 56.6      \\ \hline
			\end{tabular}
		}
		\caption{Performance comparison of different ways of transferring CLIP knowledge on HITSZ-VCM.}
		\label{table:way}
	\end{center}
\end{table}
\begin{table}[t]
	\begin{center}
		\resizebox{0.88\linewidth}{!}
		{
			\begin{tabular}{cccccccc}
				\hline
				\multicolumn{4}{c}{\multirow{2}{*}{Components}}  & \multicolumn{4}{c}{HITSZ-VCM}                     \\
				\multicolumn{4}{c}{}                     & \multicolumn{2}{c}{I2V} & \multicolumn{2}{c}{V2I} \\
				& SII        & LII        & CII        & Rank-1      & mAP       & Rank-1      & mAP       \\\hline
				1 & $\times$   & $\times$   & $\times$   & 67.7        & 57.1      & 71.4        & 56.6      \\
				2 & \checkmark & $\times$   & $\times$   & 71.5        & 58.9      & 73.6        & 57.6      \\
				3 & $\times$   & \checkmark & $\times$   & 71.2        & 59.6      & 74.0        & 58.4      \\
				4 & $\times$   & $\times$   & \checkmark & 71.0        & 59.2      & 73.8        & 58.0      \\
				5 & \checkmark & $\times$   & \checkmark & 72.0        & 59.6      & 74.5        & 58.9      \\
				6 & $\times$   & \checkmark & \checkmark & 71.8        & 60.0      & 74.3        & 58.6      \\
				7 & \checkmark & \checkmark & $\times$   & 72.9        & 59.8      & 75.5        & 59.0      \\
				8 & \checkmark & \checkmark & \checkmark & 73.4        & 60.5      & 76.1        & 59.6      \\ \hline
			\end{tabular}
		}
		\caption{Ablation studies of MII on HITSZ-VCM.}
		\label{table:mti}
		
	\end{center}
\end{table}

\textbf{More analysis about CPC.}
We further explore the impact of different ways of transferring CLIP knowledge for VVI-ReID on HITSZ-VCM in Tab.~\ref{table:way}.
The vanilla way means we directly fine-tune the visual encoder of CLIP.
As shown in Tab.~\ref{table:way}, straightforward fine-tuning or generating modality-specific textual prompts does not achieve satisfactory results in VVI-ReID.
Compared with the way of using visual memories, our proposed cross-modal update strategy brings 1.0\% mAP and 3.5\% Rank-1 gains under the V2I setting, respectively.
This is because integrating prototypes from both modalities helps the model learn modality-independent features, thereby reducing the discrepancy between modalities.
Moreover, directly conducting cross-modal collaborative learning will disrupt the network's training, resulting in decreased performance.
However, using both in collaboration can further enhance performance.
A plausible explanation is that the cross-modal update strategy allows the prototype features of two modalities to complement each other, thereby improving the stability of training.

\textbf{Effectiveness of MII.}
As shown in Tab.~\ref{table:abla1}, the proposed MII delivers a significant performance boost.
Compared with “+CPC”, incorporating MII delivers a 3.4\% gain in mAP and a 5.7\% improvement in Rank-1 on HITSZ-VCM under the I2V setting.
We believe that MII can effectively capture multi-scale temporal information within the sequence and reduce the discrepancy between modalities, thereby yielding more robust sequence-level features and contributing to enhanced performance.

\textbf{More analysis about MII.}
The proposed MII consists of three components, namely SII, LII and CII.
To verify the impact of each component, we further conduct several experiments on HITSZ-VCM, and show compared results in Tab.~\ref{table:mti}.
Ablation results indicate that each component contributes to improving cross-modality person ReID.
Removing all interaction modules leads to a clear performance drop, confirming the necessity of temporal and modality-aware feature modeling.
Introducing SII or LII individually yields noticeable gains, demonstrating their effectiveness in capturing local and global temporal patterns, respectively.
Incorporating CII alone also improves performance, validating its role in enhancing modality alignment.
\begin{figure}[t]
	\centering
	{
		\begin{tabular}{@{}c@{}c@{}}
			\includegraphics[width=0.86\linewidth]{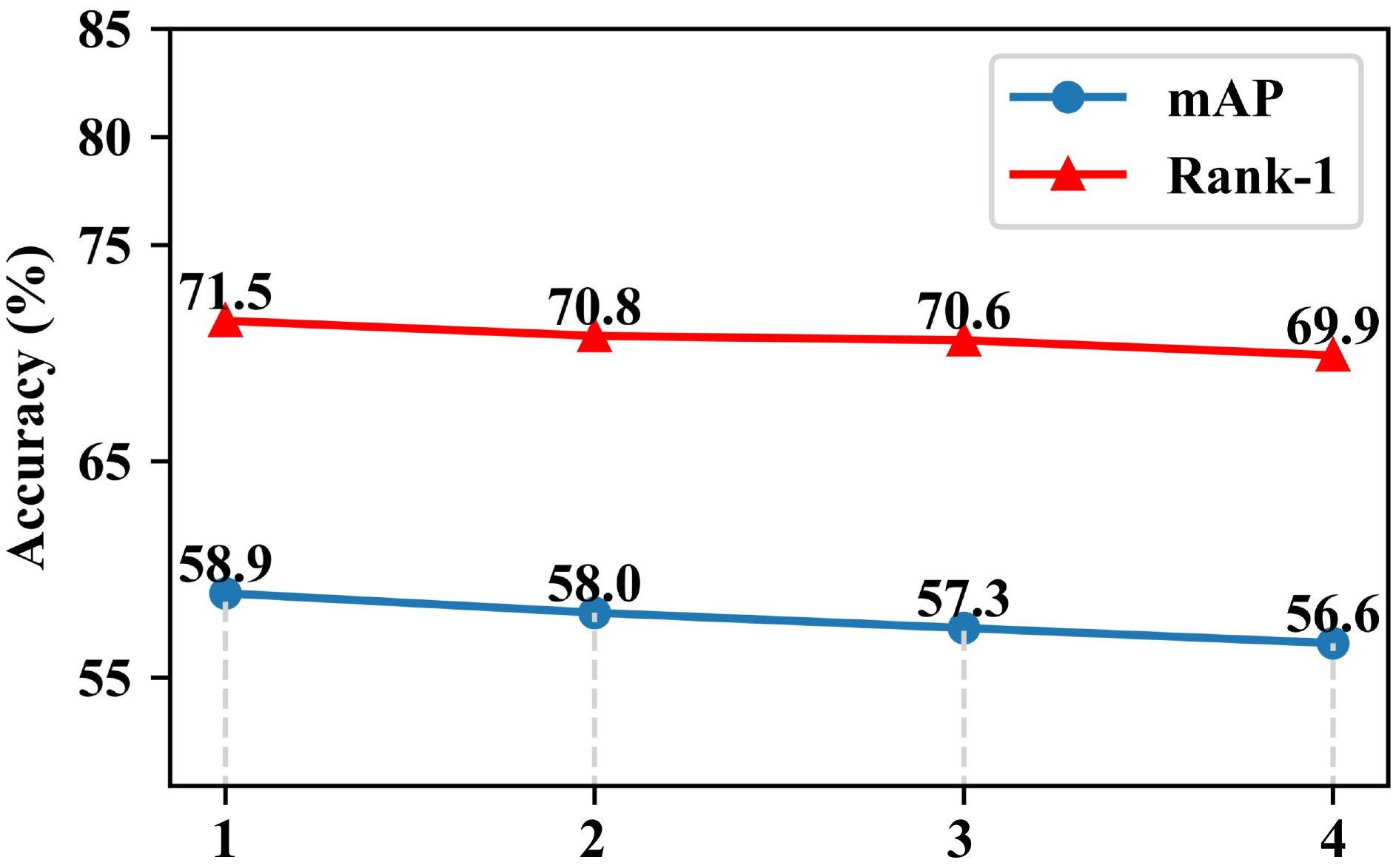}
		\end{tabular}
	}
	\caption{Illustration of the impact of time stride $S$ in SII on HITSZ-VCM under the I2V setting.
	}
	\label{fig:t1}
\end{figure}
\begin{figure}[t]
	\centering
	{
		\begin{tabular}{@{}c@{}c@{}}
			\includegraphics[width=0.88\linewidth]{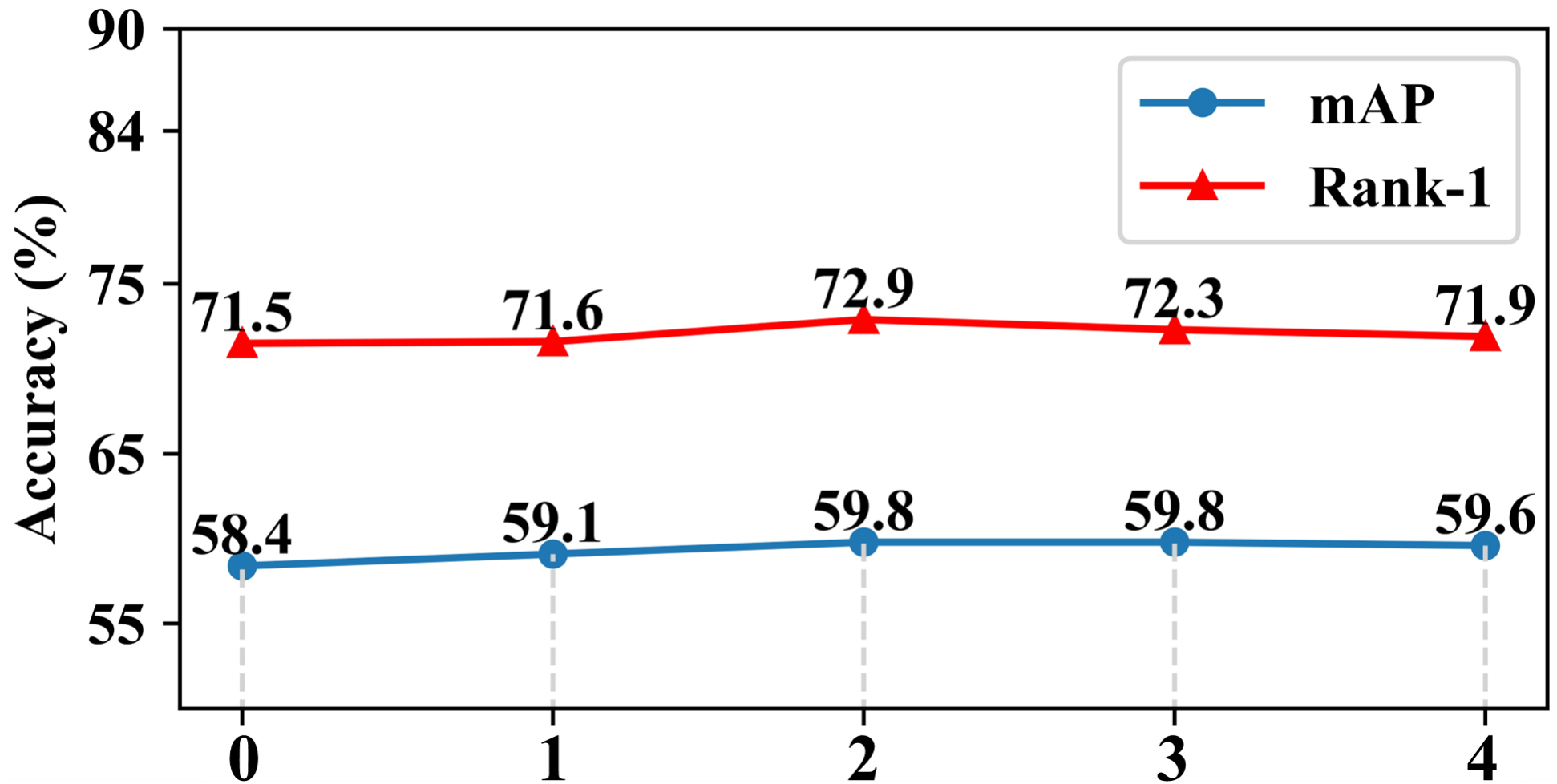}
		\end{tabular}
	}
	\caption{Illustration of the impact of time stride $S$ in LII on HITSZ-VCM under the I2V setting.
	}
	\label{fig:t2}
\end{figure}

\textbf{The impact of the time stride $S$ in MII.} The time stride $S$ in MII determines the range of temporal information captured by SII and LII.
Thus, we conduct experiments to investigate the impact of the hyper-parameter $S$ on HITSZ-VCM.
As shown in Fig.~\ref{fig:t1}, as $S$ increases from 1 to 4, the performance of SII gradually decreases.
A reasonable explanation is that exchanging patch tokens between frames that are far apart disrupts temporal consistency and local context, leading to degraded performance.
Therefore, we only perform SII in adjacent frames.
As shown in Fig.~\ref{fig:t2}, we further explore the impact of time stride $S$ in LII.
Here, $S$=0 means we do not exchange the [CLS] tokens.
As shown in Fig.~\ref{fig:t2}, if the stride is too large or too small, the performance will decrease.
As a result, we set $S$=2 in LII for our experiments.
\section{Conclusion}
In this paper, we propose a novel framework named X-ReID for VVI-ReID.
First, we propose a Cross-modal Prototype Collaboration (CPC) to effectively transfer CLIP's knowledge and reduce the modality gap.
Meanwhile, we design a Multi-granularity Information Interaction (MII) to capture multi-granular temporal information in videos and reduce the discrepancy between modalities.
Finally, information from various scales is combined to produce robust sequence-level features.
Extensive experiments show that our proposed method outperforms other state-of-the-art methods on two large-scale VVI-ReID benchmarks.
\section{Acknowledgments}
This work was supported in part by the National Natural Science Foundation of China (No.62576069, 62506272), Natural Science Foundation of Liaoning Province (No.2025-MS-025) and Dalian Science and Technology Innovation Fund (No.2023JJ11CG001).
\bibliography{aaai2026}

\begin{thebibliography}{43}
\providecommand{\natexlab}[1]{#1}

\bibitem[{Chen et~al.(2023)Chen, Zhang, Tan, Qu, and Xie}]{chen2023unveiling}
Chen, Z.; Zhang, Z.; Tan, X.; Qu, Y.; and Xie, Y. 2023.
\newblock Unveiling the power of clip in unsupervised visible-infrared person
  re-identification.
\newblock In \emph{Proceedings of the ACM International Conference on
  Multimedia}, 3667--3675.

\bibitem[{Du et~al.(2023)Du, Lei, Zhao, Dong, and Su}]{du2023video}
Du, Y.; Lei, C.; Zhao, Z.; Dong, Y.; and Su, F. 2023.
\newblock Video-based visible-infrared person re-identification with auxiliary
  samples.
\newblock \emph{IEEE Transactions on Information Forensics and Security}, 19:
  1313--1325.

\bibitem[{Feng, Wu, and Zheng(2023)}]{feng2023shape}
Feng, J.; Wu, A.; and Zheng, W.-S. 2023.
\newblock Shape-erased feature learning for visible-infrared person
  re-identification.
\newblock In \emph{Proceedings of the IEEE/CVF Conference on Computer Vision
  and Pattern Recognition}, 22752--22761.

\bibitem[{Feng et~al.(2024)Feng, Chen, Yu, Ji, Wu, Liu, Liu, Jing, and
  Luo}]{feng2024cross}
Feng, Y.; Chen, F.; Yu, J.; Ji, Y.; Wu, F.; Liu, T.; Liu, S.; Jing, X.-Y.; and
  Luo, J. 2024.
\newblock Cross-Modality Spatial-Temporal Transformer for Video-Based
  Visible-Infrared Person Re-Identification.
\newblock \emph{IEEE Transactions on Multimedia}, 26: 6582--6594.

\bibitem[{Gong et~al.(2025)Gong, Zhuge, Zhang, Yang, Zhang, and
  Lu}]{gong2025devil}
Gong, S.; Zhuge, Y.; Zhang, L.; Yang, Z.; Zhang, P.; and Lu, H. 2025.
\newblock The devil is in temporal token: High quality video reasoning
  segmentation.
\newblock In \emph{Proceedings of the IEEE/CVF Conference on Computer Vision
  and Pattern Recognition Conference}, 29183--29192.

\bibitem[{Hermans, Beyer, and Leibe(2017)}]{hermans2017defense}
Hermans, A.; Beyer, L.; and Leibe, B. 2017.
\newblock In defense of the triplet loss for person re-identification.
\newblock \emph{arXiv:1703.07737}.

\bibitem[{Hou et~al.(2024)Hou, Wang, Yan, Wu, and Xia}]{hou2024three}
Hou, W.; Wang, W.; Yan, Y.; Wu, D.; and Xia, Q. 2024.
\newblock A Three-stage Framework for Video-based Visible-Infrared Person
  Re-Identification.
\newblock \emph{IEEE Signal Processing Letters}, 31: 1254--1258.

\bibitem[{Li et~al.(2023)Li, Liu, Hu, Nie, and Yu}]{li2023intermediary}
Li, H.; Liu, M.; Hu, Z.; Nie, F.; and Yu, Z. 2023.
\newblock Intermediary-Guided Bidirectional Spatial--Temporal Aggregation
  Network for Video-Based Visible-Infrared Person Re-Identification.
\newblock \emph{IEEE Transactions on Circuits and Systems for Video
  Technology}, 33(9): 4962--4972.

\bibitem[{Li and Gong(2023)}]{li2023prototypical}
Li, J.; and Gong, X. 2023.
\newblock Prototypical contrastive learning-based CLIP fine-tuning for object
  re-identification.
\newblock \emph{arXiv preprint arXiv:2310.17218}.

\bibitem[{Li et~al.(2025)Li, Leng, Kuang, Tan, and Gao}]{li2025video}
Li, S.; Leng, J.; Kuang, C.; Tan, M.; and Gao, X. 2025.
\newblock Video-Level Language-Driven Video-Based Visible-Infrared Person
  Re-Identification.
\newblock \emph{IEEE Transactions on Information Forensics and Security},
  1--12.

\bibitem[{Li, Sun, and Li(2023)}]{li2023clip}
Li, S.; Sun, L.; and Li, Q. 2023.
\newblock CLIP-ReID: exploiting vision-language model for image
  re-identification without concrete text labels.
\newblock In \emph{Proceedings of the AAAI Conference on Artificial
  Intelligence}, volume~37, 1405--1413.

\bibitem[{Lin, Gan, and Han(2019)}]{lin2019tsm}
Lin, J.; Gan, C.; and Han, S. 2019.
\newblock Tsm: Temporal shift module for efficient video understanding.
\newblock In \emph{Proceedings of the IEEE/CVF Conference on International
  Conference on Computer Vision}, 7083--7093.

\bibitem[{Lin et~al.(2022)Lin, Li, Ma, Li, Li, Xu, Lu, and
  Zhang}]{lin2022learning}
Lin, X.; Li, J.; Ma, Z.; Li, H.; Li, S.; Xu, K.; Lu, G.; and Zhang, D. 2022.
\newblock Learning Modal-Invariant and Temporal-Memory for Video-Based
  Visible-Infrared Person Re-Identification.
\newblock In \emph{Proceedings of the IEEE/CVF Conference on Computer Vision
  and Pattern Recognition}, 20973--20982.

\bibitem[{Liu et~al.(2021)Liu, Zhang, Yu, Lu, and Yang}]{liu2021watching}
Liu, X.; Zhang, P.; Yu, C.; Lu, H.; and Yang, X. 2021.
\newblock Watching you: Global-guided reciprocal learning for video-based
  person re-identification.
\newblock In \emph{Proceedings of the IEEE/CVF Conference on Computer Vision
  and Pattern Recognition}, 13334--13343.

\bibitem[{Liu et~al.(2024)Liu, Zhang, Yu, Qian, Yang, and Lu}]{liu2024video}
Liu, X.; Zhang, P.; Yu, C.; Qian, X.; Yang, X.; and Lu, H. 2024.
\newblock A video is worth three views: Trigeminal transformers for video-based
  person re-identification.
\newblock \emph{IEEE Transactions on Intelligent Transportation Systems}.

\bibitem[{Lu, Zou, and Zhang(2023)}]{lu2023learning}
Lu, H.; Zou, X.; and Zhang, P. 2023.
\newblock Learning progressive modality-shared transformers for effective
  visible-infrared person re-identification.
\newblock In \emph{Proceedings of the AAAI Conference on Artificial
  Intelligence}, volume~37, 1835--1843.

\bibitem[{Pang et~al.(2024)Pang, Zhao, Zuo, Zhang, and Lu}]{pang2024open}
Pang, Y.; Zhao, X.; Zuo, J.; Zhang, L.; and Lu, H. 2024.
\newblock Open-vocabulary camouflaged object segmentation.
\newblock In \emph{Proceedings of the European Conference on Computer Vision},
  476--495. Springer.

\bibitem[{Park et~al.(2021)Park, Lee, Lee, and Ham}]{park2021learning}
Park, H.; Lee, S.; Lee, J.; and Ham, B. 2021.
\newblock Learning by aligning: Visible-infrared person re-identification using
  cross-modal correspondences.
\newblock In \emph{Proceedings of the IEEE/CVF International Conference on
  Computer Vision}, 12046--12055.

\bibitem[{Radford et~al.(2021)Radford, Kim, Hallacy, Ramesh, Goh, Agarwal,
  Sastry, Askell, Mishkin, Clark et~al.}]{radford2021learning}
Radford, A.; Kim, J.~W.; Hallacy, C.; Ramesh, A.; Goh, G.; Agarwal, S.; Sastry,
  G.; Askell, A.; Mishkin, P.; Clark, J.; et~al. 2021.
\newblock Learning transferable visual models from natural language
  supervision.
\newblock In \emph{International Conference on Machine Learning}, 8748--8763.
  PMLR.

\bibitem[{Schmidhuber, Hochreiter et~al.(1997)}]{schmidhuber1997long}
Schmidhuber, J.; Hochreiter, S.; et~al. 1997.
\newblock Long short-term memory.
\newblock \emph{Neural Comput}, 9(8): 1735--1780.

\bibitem[{Tian et~al.(2021)Tian, Zhang, Lin, Qu, Xie, and
  Ma}]{tian2021farewell}
Tian, X.; Zhang, Z.; Lin, S.; Qu, Y.; Xie, Y.; and Ma, L. 2021.
\newblock Farewell to mutual information: Variational distillation for
  cross-modal person re-identification.
\newblock In \emph{Proceedings of the IEEE/CVF Conference on Computer Vision
  and Pattern Recognition}, 1522--1531.

\bibitem[{Vaswani et~al.(2017)Vaswani, Shazeer, Parmar, Uszkoreit, Jones,
  Gomez, Kaiser, and Polosukhin}]{vaswani2017attention}
Vaswani, A.; Shazeer, N.; Parmar, N.; Uszkoreit, J.; Jones, L.; Gomez, A.~N.;
  Kaiser, {\L}.; and Polosukhin, I. 2017.
\newblock Attention is all you need.
\newblock In \emph{Proceedings of the Advances in Neural Information Processing
  Systems}, 5998--6008.

\bibitem[{Wang et~al.(2019)Wang, Zhang, Cheng, Liu, Yang, and
  Hou}]{wang2019rgb}
Wang, G.; Zhang, T.; Cheng, J.; Liu, S.; Yang, Y.; and Hou, Z. 2019.
\newblock RGB-infrared cross-modality person re-identification via joint pixel
  and feature alignment.
\newblock In \emph{Proceedings of the IEEE/CVF International Conference on
  Computer Vision}, 3623--3632.

\bibitem[{Wang et~al.(2025{\natexlab{a}})Wang, Gao, Niu, Zhao, and
  Feng}]{wang2025dirl}
Wang, J.; Gao, X.; Niu, S.; Zhao, H.; and Feng, G. 2025{\natexlab{a}}.
\newblock DIRL: Learning Discriminative Id-related Representations for Video
  Visible-Infrared Person Re-ID.
\newblock \emph{ACM Transactions on Multimedia Computing, Communications and
  Applications}, 1--16.

\bibitem[{Wang et~al.(2025{\natexlab{b}})Wang, Liu, Zheng, and
  Zhang}]{wang2025decoupled}
Wang, Y.; Liu, Y.; Zheng, A.; and Zhang, P. 2025{\natexlab{b}}.
\newblock Decoupled feature-based mixture of experts for multi-modal object
  re-identification.
\newblock In \emph{Proceedings of the AAAI Conference on Artificial
  Intelligence}, 8141--8149.

\bibitem[{Wang et~al.(2025{\natexlab{c}})Wang, Lv, Zhang, and
  Lu}]{wang2025idea}
Wang, Y.; Lv, Y.; Zhang, P.; and Lu, H. 2025{\natexlab{c}}.
\newblock Idea: Inverted text with cooperative deformable aggregation for
  multi-modal object re-identification.
\newblock In \emph{Proceedings of the IEEE/CVF Conference on Computer Vision
  and Pattern Recognition}, 29701--29710.

\bibitem[{Wang et~al.(2025{\natexlab{d}})Wang, Zhang, Sun, Wang, and
  Lu}]{wang2025makes}
Wang, Y.; Zhang, P.; Sun, C.; Wang, D.; and Lu, H. 2025{\natexlab{d}}.
\newblock What Makes You Unique? Attribute Prompt Composition for Object
  Re-Identification.
\newblock \emph{IEEE Transactions on Circuits and Systems for Video
  Technology}.

\bibitem[{Wu et~al.(2022)Wu, He, Liu, Yang, Lei, Mei, and Li}]{wu2022cavit}
Wu, J.; He, L.; Liu, W.; Yang, Y.; Lei, Z.; Mei, T.; and Li, S.~Z. 2022.
\newblock CAViT: Contextual alignment vision transformer for video object
  re-identification.
\newblock In \emph{Proceedings of the European Conference on Computer Vision},
  549--566. Springer.

\bibitem[{Wu et~al.(2021)Wu, Dai, Chen, Lin, Wu, Huang, Zhong, and
  Ji}]{wu2021discover}
Wu, Q.; Dai, P.; Chen, J.; Lin, C.-W.; Wu, Y.; Huang, F.; Zhong, B.; and Ji, R.
  2021.
\newblock Discover cross-modality nuances for visible-infrared person
  re-identification.
\newblock In \emph{Proceedings of the IEEE/CVF Conference on Computer Vision
  and Pattern Recognition}, 4330--4339.

\bibitem[{Yang et~al.(2022)Yang, Huang, Hu, Li, Lv, and
  Peng}]{yang2022learning}
Yang, M.; Huang, Z.; Hu, P.; Li, T.; Lv, J.; and Peng, X. 2022.
\newblock Learning with twin noisy labels for visible-infrared person
  re-identification.
\newblock In \emph{Proceedings of the IEEE/CVF Conference on Computer Vision
  and Pattern Recognition}, 14308--14317.

\bibitem[{Ye et~al.(2021{\natexlab{a}})Ye, Ruan, Du, and Shou}]{ye2021channel}
Ye, M.; Ruan, W.; Du, B.; and Shou, M.~Z. 2021{\natexlab{a}}.
\newblock Channel augmented joint learning for visible-infrared recognition.
\newblock In \emph{Proceedings of the IEEE/CVF International Conference on
  Computer Vision}, 13567--13576.

\bibitem[{Ye et~al.(2020)Ye, Shen, J.~Crandall, Shao, and Luo}]{ye2020dynamic}
Ye, M.; Shen, J.; J.~Crandall, D.; Shao, L.; and Luo, J. 2020.
\newblock Dynamic dual-attentive aggregation learning for visible-infrared
  person re-identification.
\newblock In \emph{Proceedings of the European Conference on Computer Vision},
  229--247. Springer.

\bibitem[{Ye et~al.(2021{\natexlab{b}})Ye, Shen, Lin, Xiang, Shao, and
  Hoi}]{ye2021deep}
Ye, M.; Shen, J.; Lin, G.; Xiang, T.; Shao, L.; and Hoi, S.~C.
  2021{\natexlab{b}}.
\newblock Deep learning for person re-identification: A survey and outlook.
\newblock \emph{IEEE Transactions on Pattern Analysis and Machine
  Intelligence}, 44(6): 2872--2893.

\bibitem[{Yu et~al.(2025{\natexlab{a}})Yu, Liu, Dai, Zhang, and
  Lu}]{yu2025hierarchical}
Yu, C.; Liu, X.; Dai, J.; Zhang, P.; and Lu, H. 2025{\natexlab{a}}.
\newblock Hierarchical Proxy Learning for Cloth-Changing Person
  Re-Identification.
\newblock In \emph{Proceedings of the IEEE International Conference on
  Acoustics, Speech and Signal Processing}, 1--5. IEEE.

\bibitem[{Yu et~al.(2024)Yu, Liu, Wang, Zhang, and Lu}]{yu2024tf}
Yu, C.; Liu, X.; Wang, Y.; Zhang, P.; and Lu, H. 2024.
\newblock TF-CLIP: Learning text-free CLIP for video-based person
  re-identification.
\newblock In \emph{Proceedings of the AAAI Conference on Artificial
  Intelligence}, volume~38, 6764--6772.

\bibitem[{Yu et~al.(2025{\natexlab{b}})Yu, Liu, Zhu, Wang, Zhang, and
  Lu}]{yu2025climb}
Yu, C.; Liu, X.; Zhu, J.; Wang, Y.; Zhang, P.; and Lu, H. 2025{\natexlab{b}}.
\newblock Climb-reid: A hybrid clip-mamba framework for person
  re-identification.
\newblock In \emph{Proceedings of the AAAI Conference on Artificial
  Intelligence}, 9589--9597.

\bibitem[{Yu et~al.(2025{\natexlab{c}})Yu, Dong, Zhu, Peng, and
  Tao}]{yu2025clip}
Yu, X.; Dong, N.; Zhu, L.; Peng, H.; and Tao, D. 2025{\natexlab{c}}.
\newblock Clip-driven semantic discovery network for visible-infrared person
  re-identification.
\newblock \emph{IEEE Transactions on Multimedia}.

\bibitem[{Zhang et~al.(2021)Zhang, Zhang, Qi, and Lu}]{zhang2021hat}
Zhang, G.; Zhang, P.; Qi, J.; and Lu, H. 2021.
\newblock Hat: Hierarchical aggregation transformers for person
  re-identification.
\newblock In \emph{Proceedings of the 29th ACM International Conference on
  Multimedia}, 516--525.

\bibitem[{Zhang and Wang(2023)}]{zhang2023diverse}
Zhang, Y.; and Wang, H. 2023.
\newblock Diverse embedding expansion network and low-light cross-modality
  benchmark for visible-infrared person re-identification.
\newblock In \emph{Proceedings of the IEEE/CVF Conference on Computer Vision
  and Pattern Recognition}, 2153--2162.

\bibitem[{Zhao et~al.(2025)Zhao, Pang, Yu, Zhang, Lu, Lu, Fakhri, and
  Liu}]{UniMRSeg}
Zhao, X.; Pang, Y.; Yu, C.; Zhang, L.; Lu, H.; Lu, S.; Fakhri, G.~E.; and Liu,
  X. 2025.
\newblock UniMRSeg: Unified Modality-Relax Segmentation via Hierarchical
  Self-Supervised Compensation.
\newblock In \emph{Proceedings of the Annual Conference on Neural Information
  Processing Systems}.

\bibitem[{Zhong et~al.(2020)Zhong, Zheng, Kang, Li, and Yang}]{zhong2020random}
Zhong, Z.; Zheng, L.; Kang, G.; Li, S.; and Yang, Y. 2020.
\newblock Random erasing data augmentation.
\newblock In \emph{Proceedings of the AAAI Conference on Artificial
  Intelligence}, 13001--13008.

\bibitem[{Zhou et~al.(2023)Zhou, Li, Li, Lu, Xu, and Zhang}]{zhou2023video}
Zhou, C.; Li, J.; Li, H.; Lu, G.; Xu, Y.; and Zhang, M. 2023.
\newblock Video-based visible-infrared person re-identification via style
  disturbance defense and dual interaction.
\newblock In \emph{Proceedings of ACM International Conference on Multimedia},
  46--55.

\bibitem[{Zhou et~al.(2025)Zhou, Zhou, Ren, Li, Li, and
  Lu}]{zhou2025hierarchical}
Zhou, C.; Zhou, Y.; Ren, T.; Li, H.; Li, J.; and Lu, G. 2025.
\newblock Hierarchical disturbance and Group Inference for video-based
  visible-infrared person re-identification.
\newblock \emph{Information Fusion}, 117: 102882--102900.

\end{thebibliography}

\end{document}